\documentclass[letterpaper, 10 pt, conference]{ieeeconf}  
\usepackage[utf8]{inputenc}

\IEEEoverridecommandlockouts                        

\usepackage{amsmath} 
\usepackage{amssymb}  
\usepackage{mathtools}
\usepackage{algpseudocode}
\usepackage{algorithm}
\usepackage{subcaption}
\usepackage{graphicx}
\usepackage{caption, subcaption}
\usepackage{lipsum}
\usepackage{booktabs}
\usepackage{todonotes}
\usepackage{multirow}

\makeatletter
\let\NAT@parse\undefined
\makeatother

\usepackage{array}

\newcolumntype{F}[1]{%
    >{\raggedright\arraybackslash\hspace{0pt}}p{#1}}%
\newcolumntype{T}[1]{%
    >{\centering\arraybackslash\hspace{0pt}}p{#1}}%

\usepackage[colorlinks=false, hidelinks]{hyperref}

\title{\LARGE \bf
Efficient 2D Graph SLAM for Sparse Sensing
}

\author{Hanzhi Zhou$^{*,\dag}$, Zichao Hu$^{*,\dag}$, Sihang Liu$^\dag$,  Samira Khan$^\dag$
\thanks{*The authors contributed equally.}
\thanks{\dag Department of Computer Science, University of Virginia, \{hz2zz, zh2wc, sihangliu, samirakhan\}@virginia.edu
}}
\newcommand{\norm}[1]{\left\lVert#1\right\rVert}

\DeclareMathOperator*{\argmax}{arg\,max}

\begin{document}

\maketitle
\thispagestyle{empty}
\pagestyle{empty}

\setlength{\textfloatsep}{8pt}
\begin{abstract}
Simultaneous localization and mapping (SLAM) plays a vital role in mapping unknown spaces and aiding autonomous navigation. Virtually all state-of-the-art solutions today for 2D SLAM are designed for dense and accurate sensors such as laser range-finders (LiDARs). However, these sensors are not suitable for resource-limited nano robots, which become increasingly capable and ubiquitous nowadays, and these robots tend to mount economical and low-power sensors that can only provide sparse and noisy measurements. This introduces a challenging problem called SLAM with sparse sensing. This work addresses the problem by adopting the form of the state-of-the-art graph-based SLAM pipeline with a novel frontend and an improvement for loop closing in the backend, both of which are designed to work with sparse and uncertain range data. Experiments show that the maps constructed by our algorithm have superior quality compared to prior works on sparse sensing. Furthermore, our method is capable of running in real-time on a modern PC with an average processing time of 1/100th the input interval time.
\end{abstract}

\section{INTRODUCTION}
\begin{figure}[t]
    \centering
    \begin{subfigure}[b]{0.475\columnwidth}
        \centering
        \includegraphics[width=\textwidth]{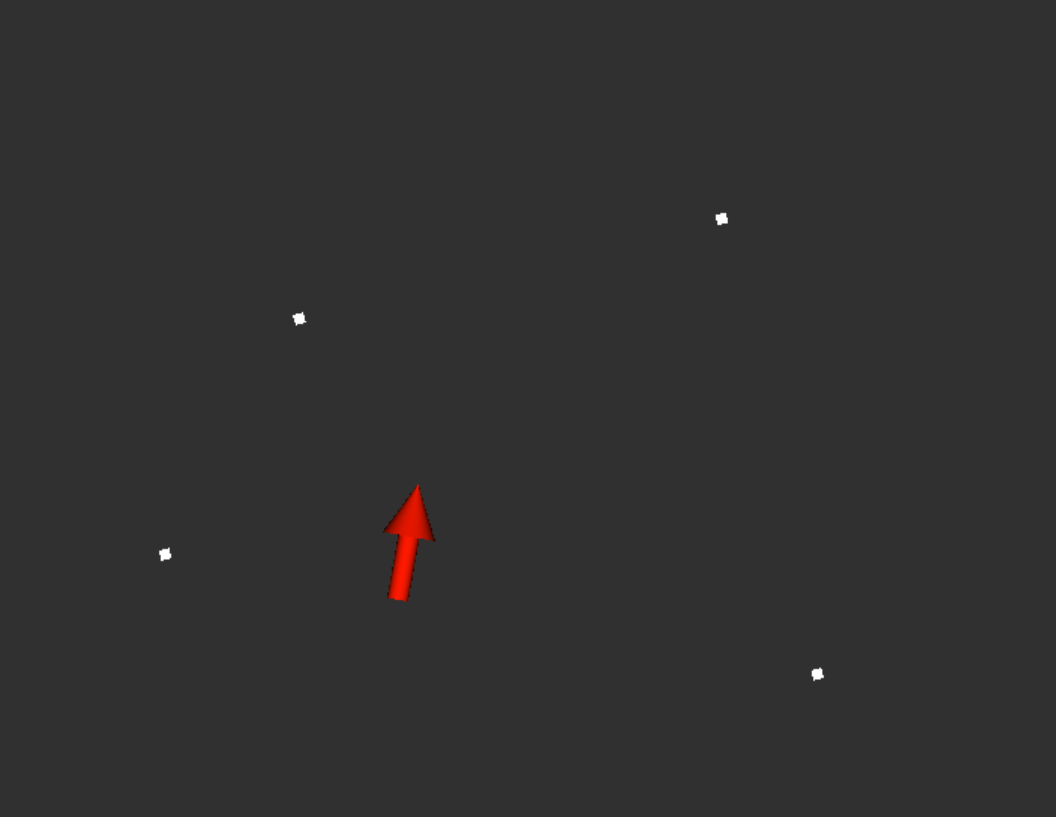}
        \caption{4 range measurements}\label{fig:sparse}
    \end{subfigure}
    \begin{subfigure}[b]{0.49\columnwidth}
        \centering
        \includegraphics[width=\textwidth]{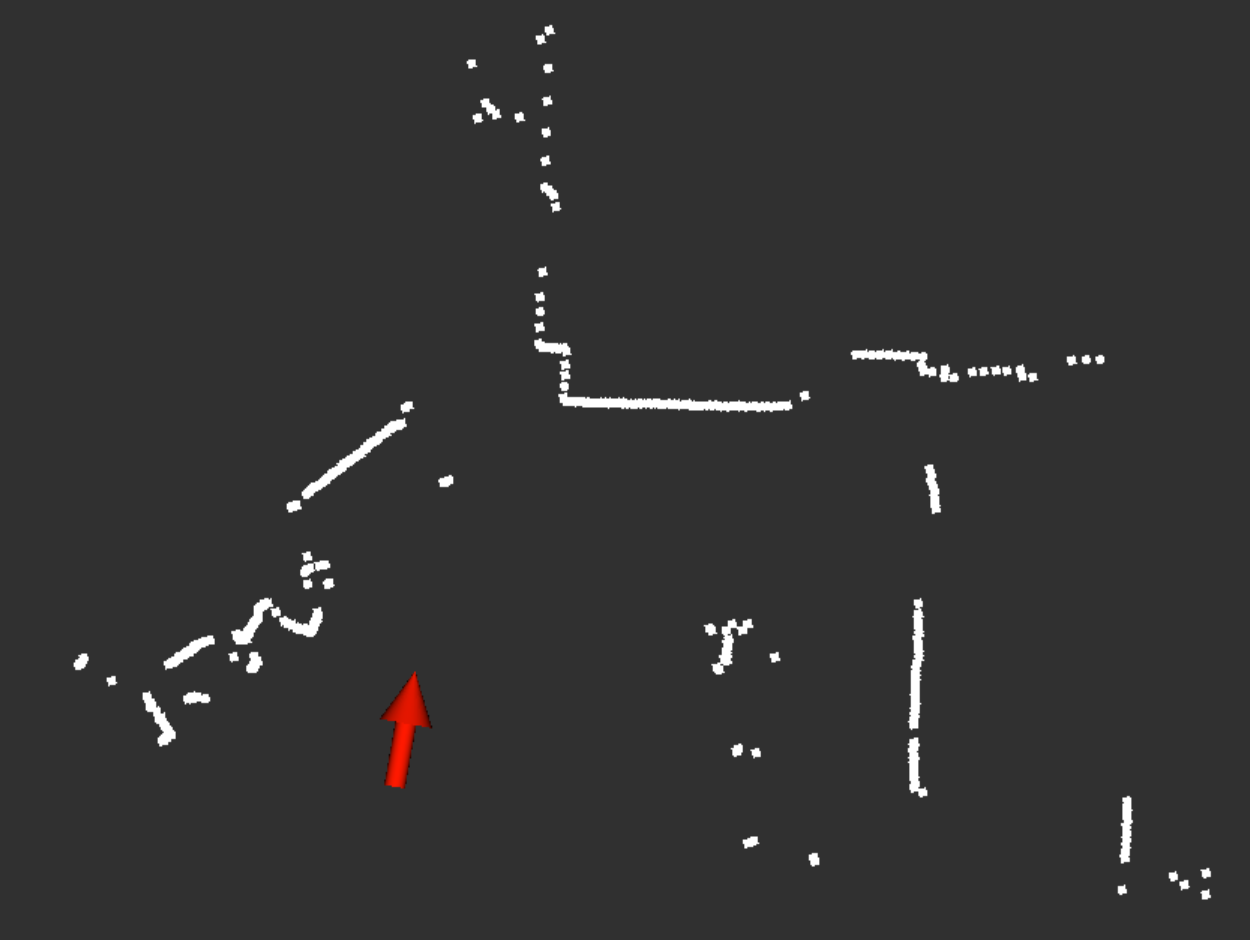}
        \caption{180 range measurements}\label{fig:dense}
    \end{subfigure}
    \caption{Sparse (left) vs dense (right) sensing}\label{fig:sparse-vs-dense}
\end{figure}

SLAM is the problem of estimating position and orientation while also constructing a map of the environment \cite{SLAM_Intro}. Solving SLAM is beneficial to navigation tasks such as path planning, robot recycling, and human decisions. Virtually all state-of-the-art 2D SLAM solutions today are designed for robots with dense and accurate sensors such as laser range-finders (LiDARs). On the contrary, recent work has shown that small, agile, and cheap nano drones demonstrate potential to carry out dangerous indoor exploration missions \cite{crazyflie1}\cite{crazyflie2}. These nano drones have limited battery and carrying capacity, and it is only possible to mount low-power sensors that can only provide sparse and noisy measurements. For example, as illustrated in Figure~\ref{fig:sparse}, the Crazyflie nano quadrotor \cite{Crazyflie} can only sense 4 range measurements\footnote{It can also sense the distances to the floor and ceiling, but they cannot contribute to 2D mapping.} at 10Hz with a maximum range of 4 meters \cite{vl53l1x}, which is almost two orders of magnitude fewer data compared to a typical 2D LiDAR with more than 180 range measurements and a range of 10 meters, as in Figure~\ref{fig:dense}. As the result, the limited sensing capacity introduces a challenging SLAM problem.

Prior work \cite{2006sparsesensing}\cite{2009sparse} has shown some progress on overcoming the SLAM with sparse sensing problem. In their work, they adopt the particle filter to solve the problem. However, particle filter has its limitations. First, it cannot refine and globally optimize the complete trajectory of the robot. Furthermore, the number of particles required for a good mapping result becomes increasingly larger when the environment space is large, slowing down computation. 

On the other hand, graph-based techniques have become the standard for modern SLAM solutions because of their superior accuracy, efficiency, and ability to refine the complete trajectory of the robot \cite{graph-vs-particle}\cite{graph-tutorial}. A specific form of a graph-based technique called pose graph optimization (PGO) has been studied the most in literature because of its simple and sparse structure, allowing it to be solved very efficiently \cite{spa}. 


PGO requires a frontend that computes accurate robot pose-to-pose relations to achieve locally consistent trajectory, typically achieved through scan-matching \cite{icp}\cite{ros_icp_scan_matcher}. It assumes that the observations from two consecutive poses have significant overlaps so that a rigid-body transformation can be calculated directly. However, sparse sensing invalidates this assumption. We propose a novel approach called landmark graph that replaces scan-matching as the frontend to address this problem. Instead of calculating a transformation by aligning observations, we utilize the fact that the sparse input accumulated over time can reflect the environment's descriptive structure (landmark). Thus, given the uncertainty of the data, we can form hypotheses of pose-to-pose and pose-to-landmark relations in a graph. Running non-linear least-square optimization updates the hypotheses and calculates a locally consistent trajectory. Our method yields similar accuracy to scan-matching, allowing PGO to be used with sparse range data. 


PGO also requires a backend that periodically establishes loop closure constraints between poses that resemble similar places. The current state-of-the-art solution, correlative scan-to-map matching \cite{corr-scan-matching}, when applied to sparse input data with high uncertainty, tends to fail frequently at differentiating the correct matches from the incorrect ones because their scores are similar. We propose an approximate match heuristic that matches each point in the scan to not a specific cell in the map but a neighborhood of cells. Our experiment shows that the heuristic makes it much easier to set a threshold to differentiate correct and incorrect matches.



To demonstrate the effectiveness of our algorithm, we perform extensive experiments on both our datasets collected with the Crazyflie nano quadrotor \cite{Crazyflie}, and Radish datasets~\cite{Radish}. For Crazyflie datasets, we show that with only 4 range measurements at 10 Hz, we can produce a map very close to the floor plan in the indoor environment. For Radish datasets, we show that our algorithm runs much faster than prior work and produces maps with superior quality. Finally, our parameter sweep experiment indicates that our algorithm can produce reasonable maps with as few as 4 measurements, whereas other mainstream 2D SLAM algorithms fail catastrophically.


To summarize, our contributions are threefold:
\begin{itemize}
    \item We are the first to provide an open-source\footnote{https://github.com/shiftlab-nanodrone/sparse-gslam} graph-based solution to solve the 2D SLAM with sparse sensing problem. Our system is capable of running in real-time on a typical modern computer.
    
    \item  We propose a novel landmark graph to replace scan-matching as the frontend for sparse range data. The landmark graph can correct relative poses and achieve locally consistent trajectories.
    
    \item We propose an approximate match heuristic to the correlative scan-to-map matching algorithm to amplify the score distinction between a correct and an incorrect match, making it easier to reject incorrect loop closures with a threshold. 
    
\end{itemize}

\section{Related Works}
Traditional SLAM solutions use filtering approaches such as Extended Kalman Filter \cite{EKF} and Particle Filter (PF) \cite{FastSLAM}\cite{FastSLAM2}. These approaches maintain poses and landmarks and perform prediction and update steps recursively. Modern works shift more attention toward the optimization approach, often known as graph-based SLAM.

First introduced by Lu and Milios in 1997 \cite{graph_slam_lu_milios}, the graph-based SLAM approaches model the SLAM problem as a sparse graph of constraints and apply nonlinear optimizations to refine robot trajectory. With the development of efficient and user-friendly backend solver frameworks \cite{g2o}\cite{isam}\cite{slam++}, graph-based approach excels in accuracy over large space, because of its ability to refine past trajectory \cite{graph_slam_lu_milios}. Additionally, advances of robust graph SLAM methods such as Switchable Constraints \cite{switch}, Dynamic Covariance Scaling (DCS) \cite{dcs}, Max-Mixture \cite{max-mixture}, and Incremental SLAM with Consistency Checking (ISCC) \cite{iscc} make graph SLAM resistant to outlier sensor measurements and improve its convergence. Many works show that graph-based SLAM can be used for a variety of sensor configurations. For example, ORB-SLAM2 \cite{orb-slam2} and RTAB-Map \cite{rtabmap} are graph-based systems that specialize in mapping with stereo and RGB-D cameras. Cartographer~\cite{cartographer} uses graph-based methods and focuses on real-time mapping using LiDARs. 

A subfield of SLAM problems is called SLAM with sparse sensing, in which the robot is limited in sensing capability and can only receive very few data points from the sensors. It is a more challenging problem because the system receives less information with more uncertainty. However, to our knowledge, only a few works have been proposed to solve this problem. Beevers et al. \cite{2006sparsesensing} group consecutive observations to extract line features as landmarks and use the Rao-Blackwellized Particle Filter (RBPF) \cite{rbpf} to solve the SLAM problem. Yap et al. \cite{2009sparse} utilizes a similar approach to tackle the problem of noisy sonar sensors, but they also applied RBPF while assuming that the walls are orthogonal to each other to produce accurate maps. However, RBPF has limitations -- it is not able to refine the past trajectory, and it gets increasingly more computation and memory intensive as the space gets larger because many more particles are required to maintain the accuracy of the map \cite{graph-vs-particle}. 

Our work aims to address the limitations of previous work on SLAM with sparse sensing. We are the first to apply a graph-based approach to this problem. Nevertheless, the adaptation of the graph-based approach is still nontrivial. Due to the lack of sufficient overlapping between different frames of the input data, conventional scan-matching techniques \cite{scan_matching_thesis} are not applicable, and the already challenging loop closure problem becomes more challenging.

\begin{figure*}[t!]
 \centering
 \vspace{5pt}
 \includegraphics[width=\textwidth]{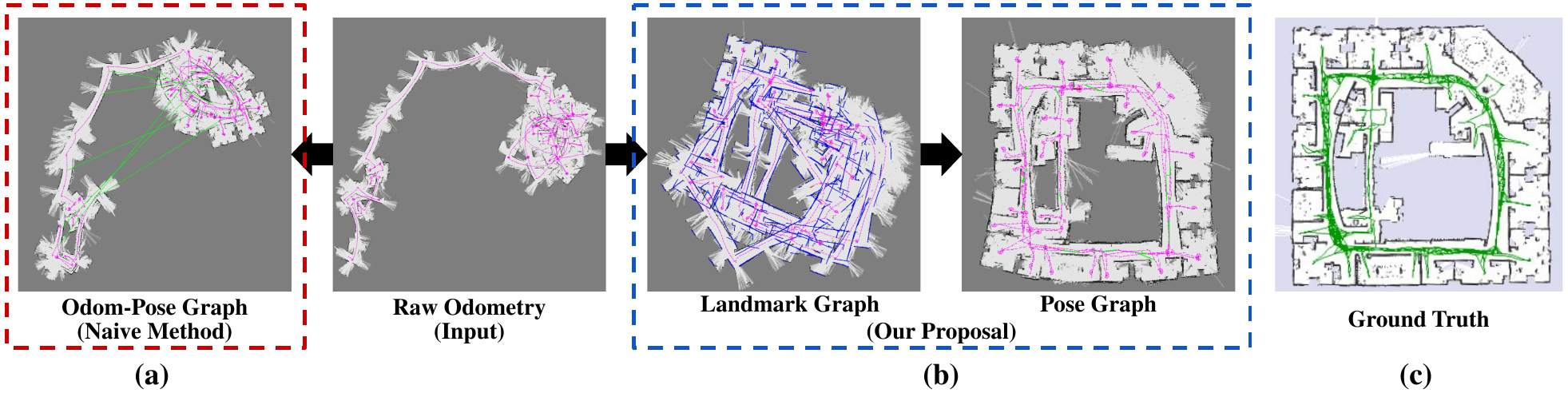}%
 \caption{Illustration of our approach on the Intel Lab dataset. For subfigures except the ground truth, the robot trajectory is in pink, landmark line segments are in blue, and loop closures are in green.}\label{fig:dual-graph}
 \vspace{-10pt}
\end{figure*}

\section{Graph SLAM with Sparse Sensing}
In order to solve SLAM with sparse sensing, we incorporate two graphs in our approach: landmark graph and pose graph. The landmark graph aims to derive accurate odometry constraints to replace scan matching as the frontend of the pose graph. We leverage the information from a group of scans collected over a period of time, known as `multiscan', and extract the line segments from the multiscan to describe the environment. Subsequently, we utilize the uncertainty (covariance) of the raw odometry and the line segments to form constraints in the landmark graph. 

As for the pose graph, we perform state copying from the landmark to obtain accurate odometry constraints and then apply correlative scan-to-map matching to perform loop closure detection periodically. We propose an approximate match heuristic to sharpen the score distinction between a good and a bad match, thus simplifying the process of finding a threshold. In the following sections, \ref{section3A} introduces the detailed formulation of the landmark graph, and \ref{section3B} presents the construction of the pose graph.


Figure~\ref{fig:dual-graph} illustrates the high-level flow of our approach and shows the need for a special frontend for sparse sensing. It can be observed that when raw odometry is noisy (e.g. not well calibrated), simply building a standard pose graph (a) from raw odometry with correct loop closures is insufficient to achieve a reasonable result. By contrast, we first constructs the landmark graph (b) to obtain a partially corrected estimate of the robot trajectory. Then, we build a pose graph by taking the landmark graph as input along with loop closures, resulting in a map close to the ground truth (c).

\subsection{Landmark Graph}\label{section3A}

Algorithm \ref{algo:landmark-update} shows the high-level procedure for updating and optimizing the landmark graph when a new measurement arrives from the sensor. The details of each line in Algorithm~\ref{algo:landmark-update} will map to the sections below.

\begin{algorithm}[t]
\caption{\strut Landmark graph update procedure}\label{algo:landmark-update}
\begin{algorithmic}[1]
\State Create a pose vertex and odometry constraint
\State Construct multiscan
\State Extract line segments from multiscan
\For {each segment in segments}
    \State Associate line segments
    \State Create a constraint and insert into graph
\EndFor
\State Save graph's state
\State Optimize landmark graph
\If{Graph is inconsistent}
    \State Remove the newly inserted constraints
    \State Restore graph's state
\Else
    \State Update each line segment's endpoints
\EndIf
\end{algorithmic}
\end{algorithm}

\subsubsection{Notation}
Let the state of the robot at time $t$ to be $\mathbf{x}_t = (x, y, \theta)^T$ and the control input to be $\mathbf{u}_t = (\Delta x, \Delta y, \Delta \theta)^T$. Then, the next robot state $\mathbf{x}_{t+1}$ can be obtained using the standard motion composition operator $\oplus$ (see section 3.2 of \cite{smith2013estimating}). The observation in the coordinate frame of $\mathbf{x}_t$ is denoted as $\mathbf{z}^t_t$, where the superscript is the index of the frame of reference and the subscript is the observation index. For 2D range measurements, $\mathbf{z}^t_t \in \mathcal{R}^{2\times n}$ represents $n$ 2D Cartesian coordinates, which we will refer as a `scan'.

\subsubsection{Multiscan construction (line 2)}
Similar to prior work \cite{2006sparsesensing}, we form a multiscan from the observations at multiple robot poses. For real-time SLAM systems, it is unreasonable to incorporate future observations, because it will cause delay in processing. Thus, we choose to construct a multiscan from $k$ previous scans: $\mathbf{z}_t, \mathbf{z}_{t-1},...,\mathbf{z}_{t-k}$. Define a transformation function $g$ on a 2d point $\mathbf{p} = (a, b)$ by a pose $\mathbf{x} = (x, y, \theta)$.
\begin{align}
    g(\mathbf{p}, \mathbf{x}) = \left[
       \begin{array}{cc}
          \cos(\theta)  & \sin(\theta) \\
          -\sin(\theta) & \cos(\theta)
       \end{array}
       \right] \mathbf{p} + \left( \begin{array}{c}
          x \\ y
       \end{array}\right)
\end{align}
In order to assemble a multiscan in the frame of $\mathbf{x}_t$, we need to transform each of the previous observations, $\mathbf{z}_{t-i}, i \in [0, k]$, to the frame of $x_t$:
\begin{equation}
    \mathbf{z}^{t}_{t-i} = g(g(\mathbf{z}^{t-i}_{t-i}, \mathbf{x}_{t-i}), \mathbf{x}_t^{-1})
\end{equation}

Suppose that $\mathbf{x}_t$ is the consequence of the robot motion $\mathbf{u}_{t-i},...,\mathbf{u}_{t-1}$. Define 
\begin{equation}
    \mathbf{u}_{it} = \mathbf{u}_{t-i} \oplus \mathbf{u}_{t-i+1} ... \oplus \mathbf{u}_{t-1}
\end{equation}
Therefore, $\mathbf{x}_t = \mathbf{x}_{t-i} \oplus \mathbf{u}_{it}$. Then, it can be shown that
\begin{align}
    \mathbf{z}^{t}_{t-i} = g(\mathbf{z}^{t-i}_{t-i}, \mathbf{u}_{it}^{-1}) \label{eq:g1}
\end{align}
We will use (\ref{eq:g1}) to transform the observations from the previous frames to the current frame.

\subsubsection{Line segment extraction (line 3)}
We implement split-and-merge to extract line segments from each multiscan, because it has the best trade-off between efficiency and accuracy as shown by Nguyen et al. \cite{line-ext-comp}. Each line is represented in polar form: $\mathbf{l} = (\rho, \alpha)$ where $\rho >= 0$ and $\alpha \in [-\pi, \pi)$ for its compactness, and we refer the reader to Garulli et al. \cite{mobile-robot-line} for the details this representation and least-square line fitting. 


\subsubsection{Line segment association (line 4-7)}
To determine if a currently observed segment is a part of an existing landmark with parameter $(\rho, \alpha)$, we use two criteria:
\begin{enumerate}
    \item The line projection error is smaller than a threshold
    \item Projected endpoints sufficiently overlap with the global line segment
\end{enumerate}

Given the endpoints $\mathbf{p}_1, \mathbf{p}_2$ of the observed segment in the global frame, criterion 1 can be formulated as
\begin{equation}\label{cri:rp}
    \norm{\mathbf{p}_1 - (\mathbf{a} + t_1 \mathbf{d})} + \norm{\mathbf{p}_2 - (\mathbf{a} + t_2 \mathbf{d})} \le \varepsilon
\end{equation}
where 
\begin{equation}\label{eq:vec-rep}
    \mathbf{a} = (\rho \cos\alpha, \rho \sin\alpha)^T,\, \mathbf{d} = (-\sin \alpha, \cos \alpha)^T
\end{equation}
\begin{equation}\label{eq:end-project}
    t_1 = (\mathbf{p}_1 - \mathbf{a}) \cdot \mathbf{d},\,\, t_2 = (\mathbf{p}_2 - \mathbf{a}) \cdot \mathbf{d}
\end{equation}
To test criterion 2, we first project the endpoints of the landmark onto itself with (\ref{eq:end-project}) to get two scalar values $s_1$ and $s_2$. Then, criterion 2 is satisfied when
\begin{equation}\label{cri:overlap}
    [s_1, s_2] \cap [t_1, t_2] \ne \emptyset
\end{equation}

For all landmarks that satisfy (\ref{cri:rp}) and (\ref{cri:overlap}), the one with the smallest projection error is associated with the current observation. If no landmarks satisfy both criteria, a new landmark is created for future associations. 

\subsubsection{Graph optimization (line 9)}
We formulate the objective function as 
\begin{align}\label{eq:lm_graph_obj}
    F(X) = \sum_i \norm{e_o(\mathbf{x}_i, \mathbf{x}_{i+1})}^2_{\Sigma_i} + \sum_{ij} \norm{e_l(\mathbf{x}_i, \mathbf{l}_{j})}^2_{\Sigma_{ij}}
\end{align}
where $e_o$ is the error function of the odometry constraints
\begin{equation}\label{eq:error-odom}
    e_o(\mathbf{x}_i, \mathbf{x}_{i+1}) = \mathbf{v}_{i,i+1}^{-1} \oplus (\mathbf{x}_i^{-1} \oplus \mathbf{x}_{i+1})
\end{equation}
where $\mathbf{v}_{i,i+1}$ is the odometry measurement between the two poses. $e_l$ is the error of the pose-landmark constraints\footnote{The error of angle $\alpha$ needs to be normalized to $[-\pi, \pi)$ range.}
\begin{equation}
    e_l(\mathbf{x}_i, \mathbf{l}_{j}) = \mathbf{v}_{ij} - f(\mathbf{x}_i^{-1}, \mathbf{l}_j)
\end{equation}
where $\mathbf{v}_{ij} = (\rho_{ij}, \alpha_{ij})$ is the measurement of the landmark $\mathbf{l}_j$ seen in the frame of $\mathbf{x}_i$, and $f(\mathbf{x}_i^{-1}, \mathbf{l}_j)$ transforms the current estimate of landmark $\mathbf{l}_j$ from the global frame to the frame of $\mathbf{x}_i$, which is given by\footnote{We need to make sure that the landmark after this transform has $\rho >=0$. If not, the angle needs to be incremented by $\pi$ and normalized again.}
\begin{equation}
    f(\mathbf{x}, \mathbf{l}) = \left( \begin{array}{c}
          \rho + x \cos(\alpha + \theta) + y \sin(\alpha + \theta) \\ \text{normAngle}(\alpha + \theta)
       \end{array}\right)
\end{equation}

To calculate covariances of odometry constraints $\Sigma_{i}$ in (\ref{eq:lm_graph_obj}), we employ first-order error propagation. Given control inputs that cause the robot to move from $\mathbf{x}_i$ to $\mathbf{x}_{i+1}$
\begin{equation}
    \mathbf{x}_{i+1} = \mathbf{x}_i \oplus \mathbf{u_1} \oplus ... \oplus \mathbf{u_n}
\end{equation}
the covariance can be calculated recursively as
\begin{align}\label{eq:odom-error-prop}
   \begin{split}
       \Sigma_{i} &= \text{Cov}(\mathbf{u}_1 \oplus ... \oplus \mathbf{u_n}) \\
       &= J_\oplus \left[
      \begin{array}{c|c}
      \text{Cov}(\mathbf{u}_1)  & \mbox{\normalfont\large\bfseries 0} \\
      \hline
      \mbox{\normalfont\large\bfseries 0} & \text{Cov}(\mathbf{u}_2 \oplus ... \oplus \mathbf{u_n})
   \end{array}
   \right] J_\oplus^T
   \end{split}
\end{align}
where $J_\oplus$ is the Jacobian of the motion composition operator $\oplus$ with respect to its inputs. To calculate the covariances of pose-landmark constraints $\Sigma_{ij}$ in (\ref{eq:lm_graph_obj}), we assume points $\mathbf{p}_k$ that constitute the given landmark observation are uncorrelated, and therefore
\begin{equation}
    \Sigma_{ij} = \sum_{k} J_{k}\,\text{Cov}(\mathbf{p}_k)\,J_{k}^T
\end{equation}
where $J_k$ is the Jacobian of the least-square line fitting function with respect to each point $\mathbf{p}_k$ whose expression is provided by Garulli et al. \cite{mobile-robot-line}. Since each point is transformed by (\ref{eq:g1}) during multiscan construction, their covariances can be approximated as
\begin{equation}
    \text{Cov}(\mathbf{p}_k) = J_g \left[
       \begin{array}{c|c}
          \text{Cov}(\mathbf{u}_{it}^{-1})  & \mbox{\normalfont\large\bfseries 0} \\
          \hline
          \mbox{\normalfont\large\bfseries 0} & \text{Cov}(\mathbf{p})
       \end{array}
       \right] J_g^T
\end{equation}
where $J_g$ is the Jacobian of $g$ w.r.t. its inputs, and
\begin{equation}
    \text{Cov}(\mathbf{u}_{it}^{-1}) = J_{\mathbf{u}_{it}^{-1}} \text{Cov}(\mathbf{u}_{it}) J_{\mathbf{u}_{it}^{-1}}^{T}
\end{equation}
where $\text{Cov}(\mathbf{u}_{it})$ can be calculated in a way similar to (\ref{eq:odom-error-prop}). $\mathbf{Cov}(\mathbf{p})$ is original source of error of the observation, which can be modeled as
\begin{equation}
    \text{Cov}(\mathbf{p}) = \sigma_d^2\left[
       \begin{array}{cc}
          \cos(\theta)^2  & \sin(\theta) \cos{\theta} \\
          \sin(\theta) \cos{\theta} & \sin(\theta)^2
       \end{array} \right]
\end{equation}
where $\theta$ is the bearing of $\mathbf{p}$ that is assumed to have no error and $\sigma_d$ is the error of this range measurement. 

After all constraints for the current observations are inserted into graph, graph optimization is performed. The optimal solution $X^* = \text{argmin} (F(X))$ of (\ref{eq:lm_graph_obj}) is solved by g2o \cite{g2o} with the Levenberg-Marquardt solver.

\subsubsection{Consistency checking (line 10-12)}
Since it is possible for line association to produce incorrect matches that introduce significant errors to the graph, we follow the idea of ISCC \cite{iscc} and implement a simplified version. Assuming the noise of the errors follows Gaussian distribution, $F(X)$ follows $\chi^2$ distribution, so we can check whether
\begin{equation}\label{eq:lm-const-check}
    F(X) \le \chi^2(0.95, n)
\end{equation}
where $\chi^2(\cdot, \cdot)$ is the inverse chi-squared CDF and $n$ is the sum of degrees of freedom of all constraints. If (\ref{eq:lm-const-check}) is not satisfied, then at least one of the constraints inserted in the current batch is an outlier. To ensure performance, they are all discarded and are not checked one by one.

\begin{figure}
\vspace{-7pt}
\end{figure}
\begin{algorithm}[t]
\caption{Line segment endpoint update procedure}\label{algo:endp}
\begin{algorithmic}[1]
\For{each landmark}
\State Initialize $s_1 = \infty$, $s_2 = -\infty$
\State Calculate vector representation $\mathbf{a}, \mathbf{d}$ with (\ref{eq:vec-rep})
\For {each observation of this landmark}
    \State Calculate $t_1$, $t_2$ of the observed segment with (\ref{eq:end-project})
    \State $s_1 = \min(s_1, t_1)$, $s_2 = \max(s_2, t_2)$
\EndFor
\State Calculate the new endpoints: $\mathbf{a} + s_1 \mathbf{d}$, $\mathbf{a} + s_2 \mathbf{d}$
\EndFor
\end{algorithmic}
\end{algorithm}

\begin{algorithm}[t!]
\caption{Pose graph update procedure}\label{algo:pose-graph}
\begin{algorithmic}[1]
\State Detect loop closures
\If{best match score $>$ threshold}
\State Copy the state of the landmark graph to pose graph
\State Prune landmark graph
\State Insert loop closure constraint with DCS kernel
\State Optimize pose graph
\EndIf
\end{algorithmic}
\end{algorithm}

\subsubsection{Line segment endpoint update (line 14)}\label{sec:endpoint-update}
The endpoints information is essential to compute (\ref{cri:overlap}). When a pose-landmark constraint is created, the endpoints of the observed segment are stored in addition to the line parameters. The algorithm to update line endpoints is shown in Algorithm \ref{algo:endp}.

\subsection{Pose Graph}\label{section3B}
The purpose of the pose graph is to produce a globally consistent map with the initial estimate from the landmark graph and the help of loop closures. The high level procedure for updating the pose graph is shown in Algorithm \ref{algo:pose-graph}.

\subsubsection{Notation}
To distinguish the vertices of the pose graph from the landmark graph, $\mathbf{y}_i$ is used to represent these vertices. Each $\mathbf{x}_i$ in the landmark graph provides initial estimate for $\mathbf{y}_i$ in the pose graph. $\mathbf{w}_{i,i+1}$ is the odometry measurement between $\mathbf{y}_i$ and $\mathbf{y}_{i+1}$, which is derived from the landmark graph as
\begin{equation}\label{eq:pose-odom-constraint}
    \mathbf{w}_{i,i+1} = \mathbf{x}_i^{-1} \oplus \mathbf{x}_{i+1}
\end{equation}

\subsubsection{Loop closure detection and approximate match heuristic (line 1-2)}


\newcommand{\round}[1]{\ensuremath{\lfloor#1\rceil}}

We mainly follow the approach of the Cartographer \cite{cartographer}. First, after every certain distance traveled, we create a local submap using the combined observations during this interval. Then, the submaps are stored as occupancy grids, in which each cell stores a probability of it being occupied. At the same time, we continuously construct multiscan from several recent poses\footnote{These poses need to have their estimates adjusted in the landmark graph before combining the sparse range measurements associated with each of them} to match against all previous submaps using correlative scan-to-map matching \cite{corr-scan-matching}, defined as finding the transformation $\mathbf{w}^*$ that best aligns the scan $h$ with submap $M$. 
\begin{equation}\label{eq:cor-sm-def}
    \mathbf{w}^* = \argmax_{\mathbf{w} \in \mathcal{W}} \sum_{i=1}^N M \round{\mathbf{w} h_i}
\end{equation}
where $\mathcal{W}$ is the search space for the transformation, $\mathbf{w} h_i$ transforms the $i$th range measurement in the scan to the submap coordinate frame, and $M\round{\cdot}$ returns the occupancy of the nearest grid cell. If the sum in (\ref{eq:cor-sm-def}) for $\mathbf{w}^*$ is greater than a threshold, we will accept the loop closure as a constraint in the pose graph. 

Nevertheless, both the multiscan and the map are noisy due to the sparsity of the input data. This frequently causes the algorithm to consider matches with points off by a few centimeters as bad matches, making it harder to differentiate good from bad matches effectively. A naive way to solve the problem is to use a larger cell size for the submap grid. However, this negatively affects the scan-to-map matching accuracy due to the decrease in submap resolution. 

To solve the problem, we use an approximate match heuristic. Instead of using the nearest cell for each point in the scan to calculate a score, we apply a 3x3 max kernel around the grid cell corresponding to each observed point.
\begin{equation}\label{eq:cor-sm-def-kernel}
    \mathbf{w}^* = \argmax_{\mathbf{w} \in \mathcal{W}} \sum_{i=1}^N \max_{x, y \in [-1, 1]} M\round{\mathbf{w} h_i + (x, y)^T}
\end{equation}

\begin{figure}[t]
    \centering
    \vspace{5pt}
    \includegraphics[width=.95\columnwidth]{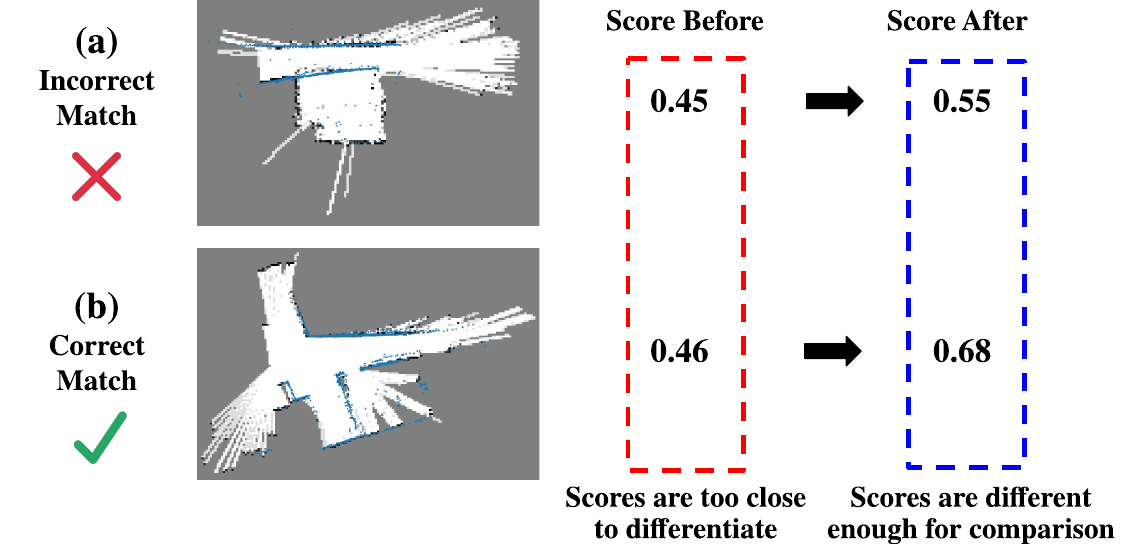}
    \caption{Blue points are observations to be matched with the submaps. Applying the max kernel makes it easier to distinguish the good match (b) from the bad match (a). }\label{fig:loop-close-match}
    \label{fig:data_association_wrong}
\end{figure}

The effect of applying this max kernel is illustrated in Figure~\ref{fig:loop-close-match}. Before applying the kernel, both the correct and incorrect matches have similar scores, making it hard to set a threshold hold to distinguish them. After applying the kernel, the score difference between them is significant enough to distinguish them effectively. 

Although (\ref{eq:cor-sm-def-kernel}) seems computational expensive due to the repeated application of the 3x3 max kernel, in reality, one can precompute the submap occupancy grid with the kernel applied because submaps are fixed after construction. Therefore, our method (\ref{eq:cor-sm-def-kernel}) have no performance penalty compared to the original method (\ref{eq:cor-sm-def}) besides a small submap initialization cost.  

\subsubsection{Graph state copying and pruning (line 3-4)}
To make use of the optimized pose estimates in the landmark graph, the odometry constraints between consecutive pose vertices are calculated with (\ref{eq:pose-odom-constraint}). Additionally, the estimates of the poses in the pose graph are copied from the landmark graph after each successful loop closure. Note that only the estimates of the poses inserted after the last loop closure optimization are copied. This ensures that we keep the estimates of vertices that are already optimized. 

After the state is copied, the landmark graph can be pruned by fixing the recently added vertices and edges and removing all other edges and constraints. This pruning procedure ensures that the landmark graph's size remains small, thus making it efficient even in prolonged exploration tasks. 

\subsubsection{Graph optimization (line 4-5)}
The pose graph objective follows the classical formulation as introduced by Sünderhauf et al. \cite{pose-graph-for}, which is given by
\begin{equation}\label{eq:pose-obj}
    F(X) = \sum_i \norm{e_o(\mathbf{y}_i, \mathbf{y}_{i+1})}^2_{\Sigma_i} + \sum_{ij} \norm{e_{lc}(\mathbf{y}_i, \mathbf{y}_{j})}^2_{\Sigma_{ij}}
\end{equation}
where $e_o$ is identical to (\ref{eq:error-odom}), and 
\begin{equation}
    e_{lc}(\mathbf{y}_i, \mathbf{y}_{j}) = \mathbf{w}_{ij}^{-1} \oplus (\mathbf{y}_i^{-1} \oplus \mathbf{y}_j)
\end{equation}
where $\mathbf{w}_{ij}$ represents a rigid transformation between pose $\mathbf{y}_i$ and $\mathbf{y}_j$ calculated by the loop closure detector. In (\ref{eq:pose-obj}), the covariance matrices for odometry constraints, $\Sigma_i$, are copied from the landmark graph. The covariance matrices for loop closure constraints, $\Sigma_{ij}$, are calculated by fitting a Gaussian distribution to the neighborhood of $\mathbf{w}_{ij}$ as in Olson \cite{corr-scan-matching}. 

However, the least-square formulation of graph SLAM by itself is not resistant to outlier constraints, which may arise due to uncertainty and ambiguity of measurement and the environment. Sparse sensing makes this problem worse by providing less measurement with more uncertainty. Hence, robust SLAM methods are critical to ensure the success of our algorithm. A few robust SLAM methods can work with our model, such as ISCC \cite{iscc}, and DCS \cite{dcs}. As per our experiments, DCS works the best in our implementation, and it is fairly easy to tune. Finally, we used Gauss-Newton algorithm provided by g2o \cite{g2o} to minimize (\ref{eq:pose-obj}).

\section{Experiments}\label{section4}

To demonstrate the effectiveness of our algorithm, we perform extensive experiments on the well-established Radish \cite{Radish} datasets and real-world sparse sensing datasets we collected. The quality of maps are measured by \textit{absolute translational error} and \textit{absolute rotational error}, proposed by K\"{u}mmerle et al. \cite{measure_metric} who also provide the ground truth for the datasets. When not otherwise specified, the grid cell size for our maps used in the experiment is 0.1m. 4 range measurements are sampled from each scan to simulate sparsity, and each range measurement is capped at 5m. Multiscan size is 120, meaning that 30 consecutive 4-point scans are grouped as the input for the landmark graph.

\subsection{Comparison among approximate matching kernels}
\begin{figure}[t]
    \centering
    \vspace{5pt}
    \includegraphics[width=\columnwidth]{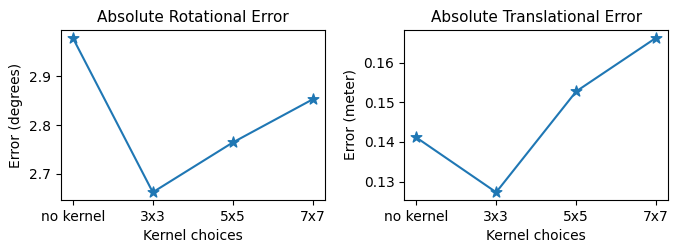}%
    \caption{Kernel choices vs mapping error}\label{fig:kernel}
    \vspace{-10pt}
\end{figure}

To justify the use of the 3x3 kernel in (\ref{eq:cor-sm-def-kernel}), we compare the quality of the maps constructed from the Intel Lab dataset using no kernel, 3x3 kernel, 5x5 kernel, and 7x7 kernel. As shown in Figure~\ref{fig:kernel}, 3x3 kernel is the ideal size that provides the appropriate fuzziness against sparse and noisy data, yielding the lowest mapping error. Using larger kernel sizes will harm matching accuracy due to the significantly increased tolerance for mismatches.

\subsection{Comparison among multiscan sizes}
\begin{figure}[t]
    \centering
    \includegraphics[width=\columnwidth]{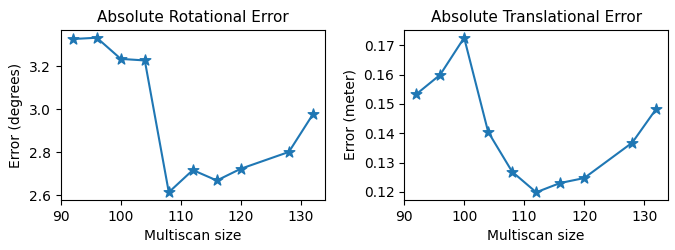}%
    \caption{Multiscan sizes vs mapping error}\label{fig:multiscan_exp}
\end{figure}

For our proposed landmark graph to work, an appropriate multiscan size needs to be chosen. As shown in Figure~\ref{fig:multiscan_exp}, although our algorithm can work with a wide range of multiscan sizes, there is a "sweet spot" in the middle which gives low errors. The reason is that when the multiscan size is too small, fewer line segments can be extracted, causing the frontend's accuracy to degrade. On the other hand, when the multiscan size is too large, errors from odometry will accumulate more, causing the extracted line segments to tilt away from the ground truth. The middle range provides good balances between data density and error accumulation. 

\subsection{Sensitivity study on the number of range measurements}
\begin{figure}[t]
    \centering
    \vspace{5pt}
    \includegraphics[width=\columnwidth]{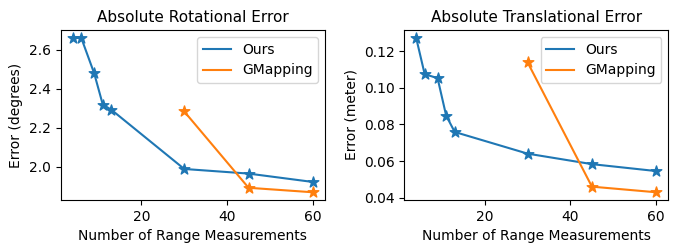}%
    \caption{Number of range measurements vs mapping error}\label{fig:sweep}
\end{figure}

Although our algorithm is designed for sparse sensing, it will be interesting to see how the algorithm performs as measurements get denser. In Figure~\ref{fig:sweep}, we vary the number of range measurements sampled from the Intel Lab dataset from 4 to 60 (out of 180). For reference, we also present the results of GMapping \cite{gmapping}.\footnote{We also experimented Hector SLAM \cite{hector} and Cartographer \cite{cartographer}. However, as they are designed for dense sensing (e.g., via modern LiDARs), they do not produce meaningful output under moderately sparse settings (e.g., less than 60 range measurements).}


We can observe that the mapping accuracy improves with the addition of new range measurements. While GMapping fails under 30 range measurements, our method can still produce reasonable maps, even when there are as few as 4 range measurements per scan. 

Note that when measurements get denser, diminishing return can be observed. The explanation for diminishing return is that our frontend is tailored for sparse data by interpolating probable line segments, and it does not benefit directly and significantly from the additional characteristics exhibited by the increasing density of data. 

\begin{figure*}[t!]
    \centering
    \vspace{5pt}
    \begin{subfigure}[b]{0.33\textwidth}
        \centering
        \includegraphics[width=0.98\textwidth]{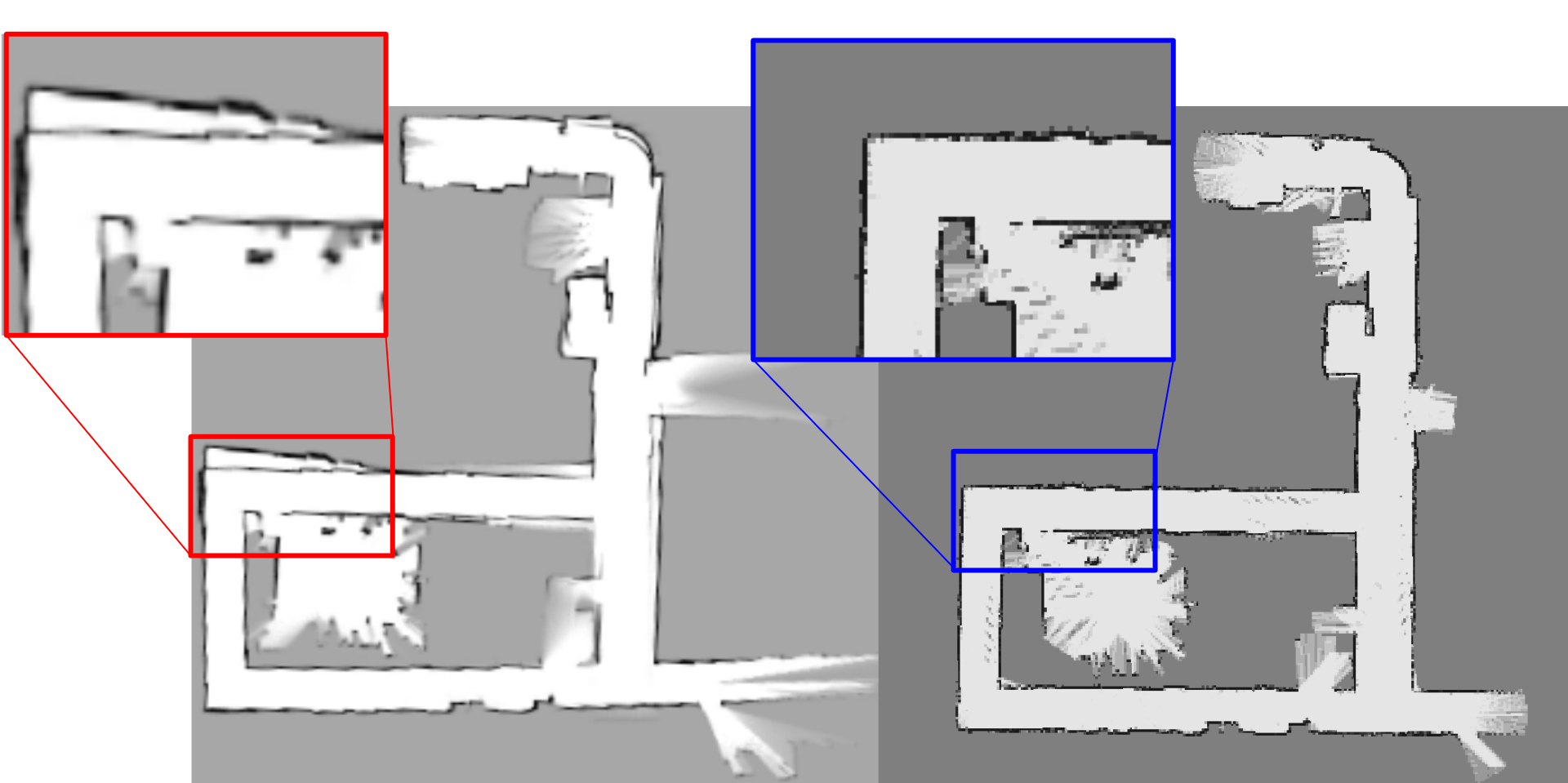}
        \caption{CMU NSH}
        \label{fig:cmu-nsh}
    \end{subfigure}%
    \begin{subfigure}[b]{0.33\textwidth}
        \centering
        \includegraphics[width=0.98\textwidth]{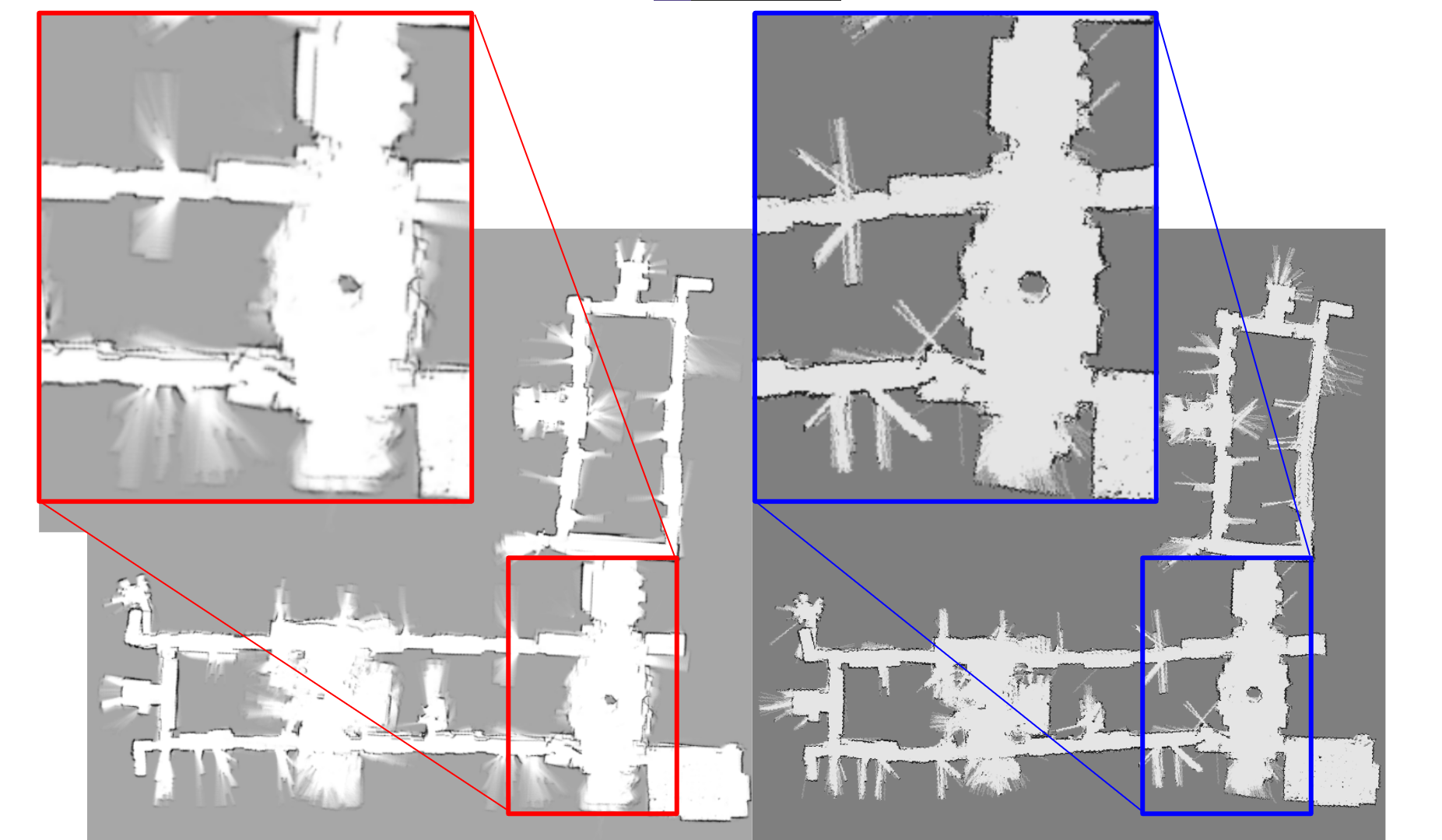}
        \caption{Stanford Gates}
        \label{fig:stanford}
    \end{subfigure}%
    \begin{subfigure}[b]{0.33\textwidth}
        \centering
        \includegraphics[width=0.98\textwidth]{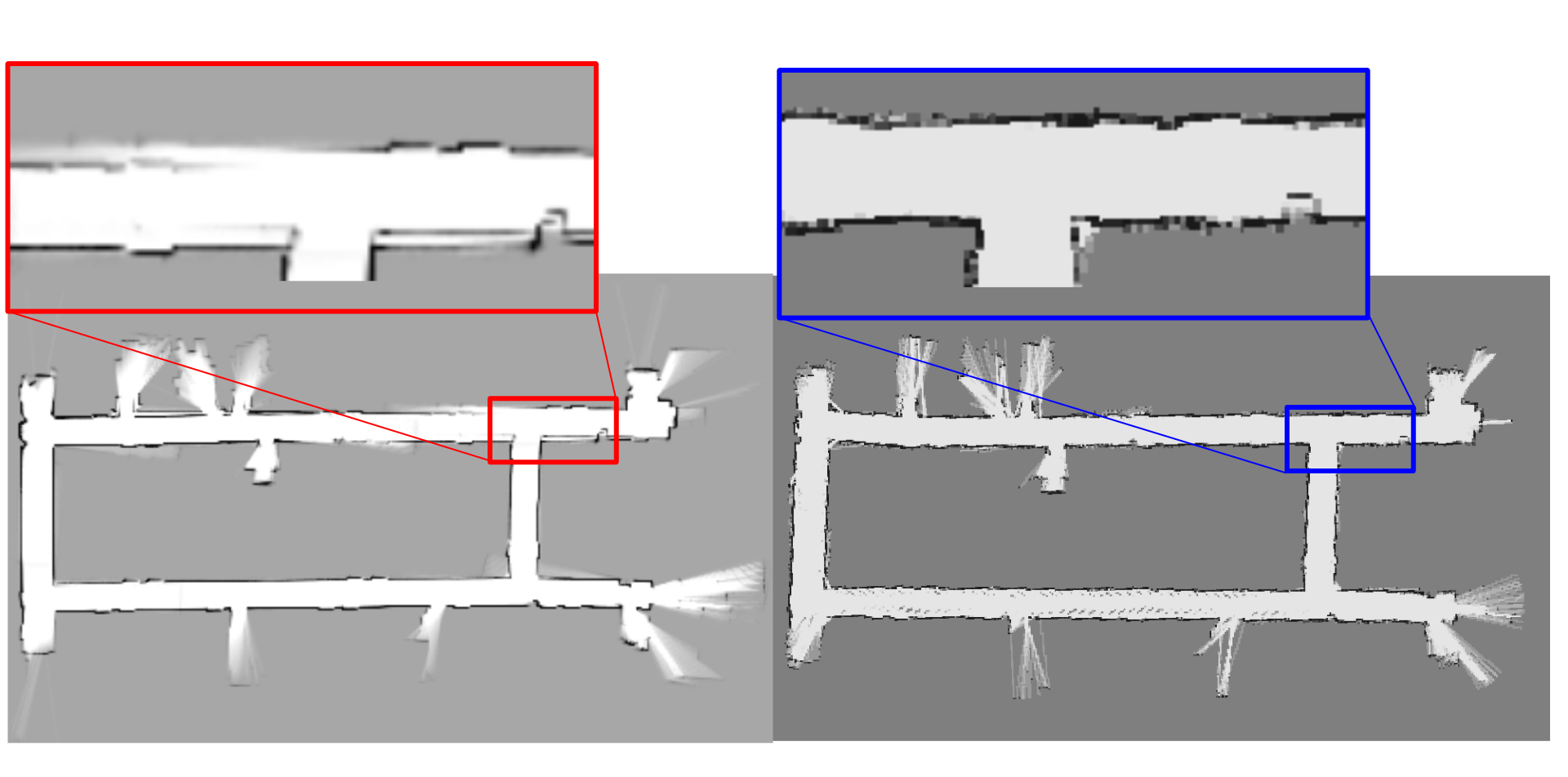}
        \caption{USC SAL}
        \label{fig:usc-sal}
    \end{subfigure}
    \caption{For each subfigure, the left image is from prior work \cite{2006sparsesensing} while the right image is our result. We highlight the visual details to demonstrate the better quality of our generated maps.}\label{fig:2006-comp}
    \vspace{-10pt}
\end{figure*}


\subsection{Comparison with previous works}\label{radish evaluation}
We first run our algorithm on the datasets used by Beevers et al. \cite{2006sparsesensing} We follow their way to simulate sparse sensing -- sub-sample range measurements and set distance cap -- for a fair comparison. As shown in Figure~\ref{fig:2006-comp}, our algorithm has more solid and clear representation of the walls. Furthermore, unlike previous work, our algorithm is less susceptible to spurious landmarks. For example, we highlight a few places in Figure~\ref{fig:2006-comp} where the previous work has noticeable aliasing while we do not. 

Second, we evaluate our approach on Aces, Intel Lab, and MIT Killian datasets from Radish. Even for a SLAM system with dense input data, it is challenging to produce good results on these datasets because (1) Intel Lab's odometry is extremely noisy, (2) Aces lacks revisits of the same places, which makes loop closure more challenging, and (3) MIT Killian dataset has a considerable dimension of roughly 190m by 240m. Since no quantitative baselines for sparse sensing exist, we have no choice but to compare our results with GMapping \cite{gmapping}, a dense sensing method. 

In Table~\ref{tb:metric} we present results of GMapping with 30 range measurements (30pt), which is the lowest number that can yield a good map for each dataset. While our method can work with as few as 4 range measurements, we present results using 11 range measurements (11pt), which can produce results as least comparable to the results of GMapping across all datasets. As we can observe, our proposal can produce results comparable to or better than GMapping with much fewer range measurements. Additionally, GMapping fails to complete the MIT Killian dataset within a reasonable amount of time due to the large dataset size and high number of particles needed to produce good maps under sparse settings. 




\begin{table}[t]
\centering
\caption{Quantitative comparison with GMapping}\label{tb:metric}
\begin{tabular}{@{}ccc@{}}
\toprule
                                                 & \multicolumn{1}{c}{This Work}               & \multicolumn{1}{c}{GMapping}          \\ \midrule
\multicolumn{1}{l}{Aces}                         & \multicolumn{1}{c}{11pt}                & \multicolumn{1}{c}{30pt} \\
\multicolumn{1}{l}{\quad Absolute translational} & \multicolumn{1}{c}{$0.0455 \pm 0.0492$}  & \multicolumn{1}{c}{$0.1040 \pm 0.2839$}   \\
\multicolumn{1}{l}{\quad Absolute rotational}    & \multicolumn{1}{c}{$1.159 \pm 1.440$}  & \multicolumn{1}{c}{$1.334 \pm 2.421$}   \\
\midrule
\multicolumn{1}{l}{Intel Lab}                    & \multicolumn{1}{c}{11pt}               & \multicolumn{1}{c}{30pt} \\
\multicolumn{1}{l}{\quad Absolute translational} & \multicolumn{1}{c}{$0.0848 \pm 0.1151$}  & \multicolumn{1}{c}{$0.1139 \pm 0.2274$}   \\
\multicolumn{1}{l}{\quad Absolute rotational}    & \multicolumn{1}{c}{$2.319 \pm 2.371$}  & \multicolumn{1}{c}{$2.283 \pm 2.331$}   \\
\midrule
\multicolumn{1}{l}{MIT Killian}                  & \multicolumn{1}{c}{11pt}               & \multirow{3}{*}{Fail to produce map} \\
\multicolumn{1}{l}{\quad Absolute translational} & \multicolumn{1}{c}{$0.0718 \pm 0.1913$}  &    \\
\multicolumn{1}{l}{\quad Absolute rotational}    & \multicolumn{1}{c}{$2.065 \pm 3.846$}  &    \\
\bottomrule
\end{tabular}
\end{table}

\subsection{Real world sparse sensing datasets}\label{crazyflie evaluation}
\begin{figure}[t]
    \centering
    \vspace{5pt}
    \includegraphics[width=\columnwidth]{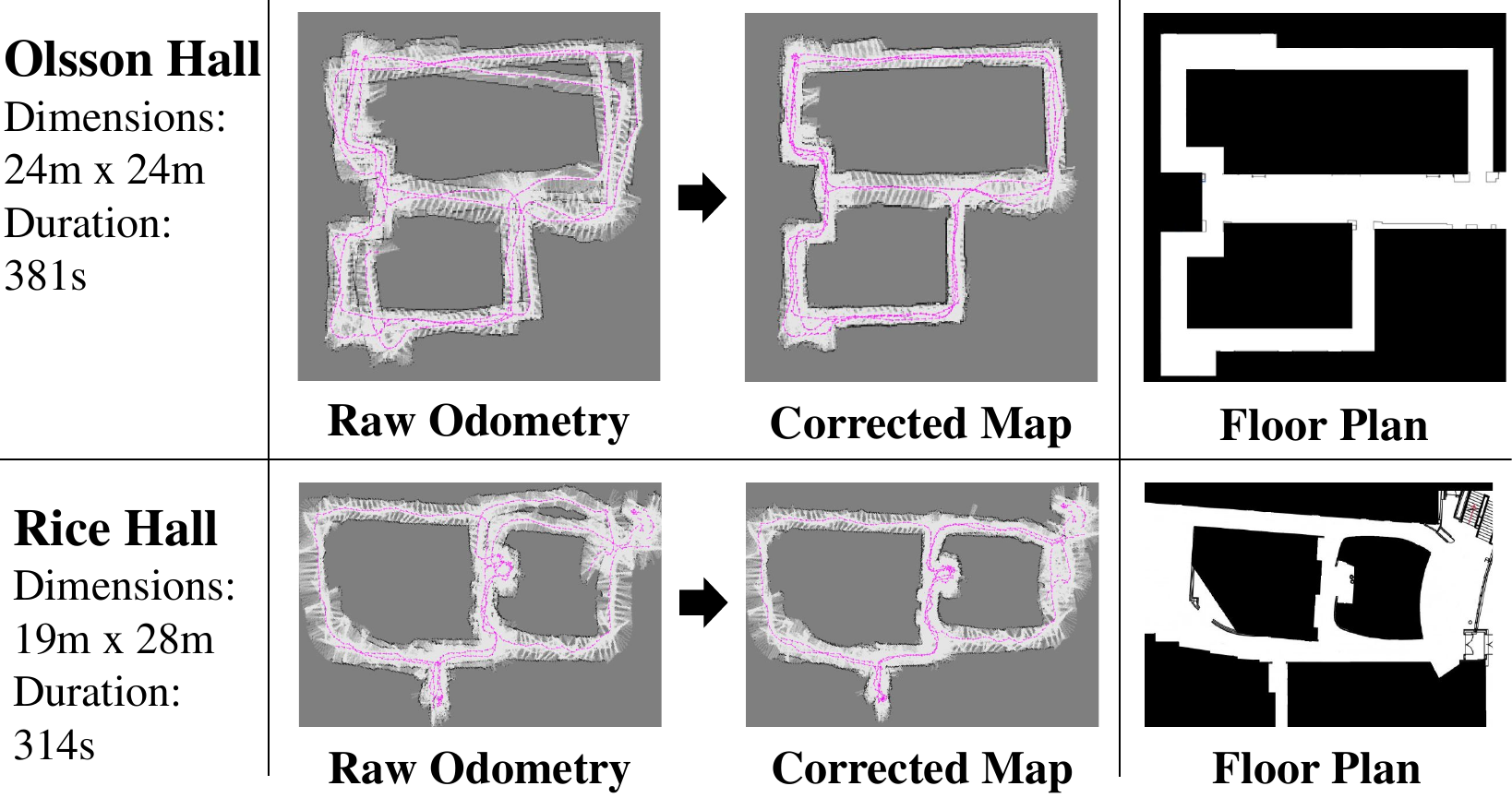}%
    \caption{Maps built with data from Crazyflie compared with the floor plan. Drone trajectory is in pink. The irrelevant details of the floor plan are covered in black.}\label{fig:crazyflie experiment}
\end{figure}

Besides using sub-sampling to simulate sparse sensing, we demonstrate our work on real-world robots with sparse range sensors. We utilized the Crazyflie \cite{Crazyflie} nano-quadrotor, which weighs only 27g and is capable of estimating its trajectory with its IMU and PMW3901 optical flow sensor. We equip it with 4 VL53L1x ToF sensors providing distance to the front, back, left, and right, with an effective range of 2m and at a rate of 10Hz. We collect around 5 to 6 minutes of flight data in Olsson Hall and Rice Hall of University of Virginia by driving the drone around manually. As shown in Figure \ref{fig:crazyflie experiment}, although the drone can localize itself with dead-reckoning to some extent, the odometry error still accumulates over time, leading to map aliasing. Our algorithm can correct the map and produce results similar to the floor plan. Since the range measurements are incredibly sparse, it is worth noting that some regions of the occupancy grid are not completely filled. 

\subsection{Speed Evaluation}\label{time evaluation}
\begin{table}[t]
\vspace{-5pt}
\centering
\caption{Speed evaluation. Unit is in seconds}\label{tb:perf}
\begin{tabular}{@{}lllll@{}}
\toprule
Dataset                         & ACES  & Intel Lab & MIT Killian\\\midrule
\textbf{Average data interval}             & \textbf{0.185}  & \textbf{0.197}     & \textbf{0.439}\\
Mean processing time        & 0.0008 & 0.0022         & 0.0061\\
Max frontend processing time    & 0.011  & 0.011         & 0.021\\
Max backend processing time     & 0.106     & 0.214         & 0.654\\
GMapping mean processing time        & 0.348   & 0.240      & Fail \\
\bottomrule
\end{tabular}
\end{table}

\begin{figure}[t]
    \centering
    \begin{subfigure}[b]{0.48\columnwidth}
        \centering
        \includegraphics[width=\textwidth]{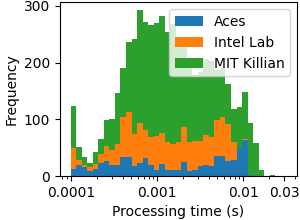}
        \caption{Frontend}\label{fig:frontend-t}
    \end{subfigure}
    \begin{subfigure}[b]{0.48\columnwidth}
        \centering
        \includegraphics[width=\textwidth]{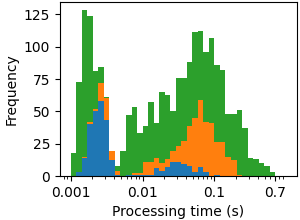}
        \caption{Backend}\label{fig:backend-t}
    \end{subfigure}
    \caption{Processing time distribution. Time is in log scale.}\label{fig:time-eval}
    \vspace{-5pt}
\end{figure}

To demonstrate the real-time capability of our algorithm, we evaluate our system's processing time on a Desktop PC with Intel Core i7-9700K. The frontend (Algorithm~\ref{algo:landmark-update}) runs synchronously with the input data, while the backend (Algorithm~\ref{algo:pose-graph}) runs asynchronously. We summarize the key statistics in Table~\ref{tb:perf} and the time distribution as a stacked histogram in Figure~\ref{fig:time-eval}. It can be observed that our method use only 1/100th of the data interval on average with most processing happens very quickly. This can be attributed to the efficient frontend and graph pruning. Occasional processing time peaks in frontend are also well below the average data interval. For the backend, its processing time allows it to execute a few times per second in the background, allowing loops to be closed very promptly. 

On the other hand, GMapping spends significant amount of time for each dataset, due to need for many particles for map to be reasonably accurate. This shows why graph-based approaches are more suited for sparse sensing than particle based methods used in prior works. 

\section{Conclusions}
This paper presented the first graph-based system to address challenging SLAM with sparse sensing problems. The solution incorporated a novel frontend analogue to scan matching but tailored for sparse sensing and an improved loop closure detection algorithm. Our system is evaluated using various datasets, and it shows promising ability to handle even large real-world indoor exploration tasks. Possible future directions of research include extending the algorithm to solve the multi-robot SLAM with sparse sensing. 




\section*{ACKNOWLEDGMENT}
Radish datasets \cite{Radish} are used to benchmark our algorithm. Thank Patrick Beeson, Dirk Hähnel, Mike Bosse, John Leonard, Andrew Howard, Nick Roy, and Brian Gerkey for providing these datasets.

\bibliographystyle{IEEEtran}
\bibliography{references.bib}

\end{document}